\documentclass{article}

\usepackage{microtype}
\usepackage{graphicx}
\usepackage{subfigure}
\usepackage{booktabs} 
\usepackage{colortbl}
\usepackage[table]{xcolor}
\usepackage{hyperref}
\definecolor{lightgray}{gray}{0.92}
\definecolor{bestgreen}{RGB}{220,245,220}

\usepackage[accepted]{icml2025}

\usepackage{amsmath}
\usepackage{amssymb}
\usepackage{mathtools}
\usepackage{amsthm}

\usepackage[capitalize,noabbrev]{cleveref}

\theoremstyle{plain}

\theoremstyle{definition}

\theoremstyle{remark}

\usepackage[textsize=tiny]{todonotes}

\icmltitlerunning{Multi-Anchor Guided Diffusion}

\begin{document}

\twocolumn[
\icmltitle{MAP-Diff: Multi-Anchor Guided Diffusion for Progressive 3D Whole-Body Low-Dose PET Denoising}



\icmlsetsymbol{equal}{$\dagger$}

\begin{icmlauthorlist}
\icmlauthor{Peiyuan Jing}{ZHAW,IC}
\icmlauthor{Chun-Wun Cheng}{Cam}
\icmlauthor{Liutao Yang}{IC}
\icmlauthor{Zhenxuan Zhang}{IC}
\icmlauthor{Thiago V. Lima}{Luc,ZHAW}
\icmlauthor{Klaus Strobel}{Luc}
\icmlauthor{Antoine Leimgruber}{Luc}
\icmlauthor{Angelica Aviles-Rivero}{Tsu}
\icmlauthor{Guang Yang}{equal,IC,nhl,Crc,Kcl}
\icmlauthor{Javier A. Montoya-Zegarra}{equal,ZHAW,Luc}

\end{icmlauthorlist}

\icmlaffiliation{ZHAW}{School of Engineering, Zurich University of Applied Sciences, CH}
\icmlaffiliation{IC}{Bioengineering Department and Imperial-X, Imperial College London, UK}
\icmlaffiliation{Cam}{DAMTP, University of Cambridge, UK}
\icmlaffiliation{Luc}{Lucerne University Teaching and Research Hospital, CH}
\icmlaffiliation{nhl}{National Heart and Lung Institute, Imperial College London, UK}
\icmlaffiliation{Crc}{Cardiovascular Research Centre, Royal Brompton Hospital, UK}
\icmlaffiliation{Kcl}{School of Biomedical Engineering \& Imaging Sciences, King's College London, UK}
\icmlaffiliation{Tsu}{Yau Mathematical Sciences Center, Tsinghua University, CN}

\icmlcorrespondingauthor{Javier Montoya}{javier.montoya@zhaw.ch}


\vskip 0.3in]



\printAffiliationsAndNotice{$\dagger$ co last authors} 

\begin{abstract}
Low-dose Positron Emission Tomography (PET) reduces radiation exposure but suffers from severe noise and quantitative degradation. Diffusion-based denoising models achieve strong final reconstructions, yet their reverse trajectories are typically unconstrained and not aligned with the progressive nature of PET dose formation.
We propose MAP-Diff, a multi-anchor guided diffusion framework for progressive 3D whole-body PET denoising. MAP-Diff introduces clinically observed intermediate-dose scans as trajectory anchors and enforces timestep-dependent supervision to regularize the reverse process toward dose-aligned intermediate states. Anchor timesteps are calibrated via degradation matching between simulated diffusion corruption and real multi-dose PET pairs, and a timestep-weighted anchor loss stabilizes stage-wise learning. At inference, the model requires only ultra-low-dose input  while enabling progressive, dose-consistent intermediate restoration. 
Experiments on internal (Siemens Biograph Vision Quadra) and cross-scanner (United Imaging uEXPLORER) datasets show consistent improvements over strong CNN-, Transformer-, GAN-, and diffusion-based baselines. On the internal dataset, MAP-Diff improves PSNR from 42.48 dB to 43.71 dB (+1.23 dB), increases SSIM to 0.986, and reduces NMAE from 0.115 to 0.103 (-0.012) compared to 3D DDPM. Performance gains generalize across scanners, achieving 34.42 dB PSNR and 0.141 NMAE on the external cohort, outperforming all competing methods.

\end{abstract}

\section{Introduction}\label{sec:intro}
Positron Emission Tomography (PET) plays a central role in clinical diagnosis, treatment planning, and quantitative functional analysis~\cite{czernin2002positron}. 
However, PET acquisition requires the injection of radioactive tracers, which inevitably introduces radiation exposure to patients~\cite{berger2003positron}. 
To reduce this risk, clinical protocols often lower the injected dose or shorten the acquisition time, but both strategies substantially degrade image quality due to increased noise and reduced count statistics~\cite{fletcher2008recommendations,wang2021artificial}. 
This creates a fundamental trade-off between radiation safety and diagnostic reliability. 
As a result, developing robust low-dose PET denoising methods that can recover high-fidelity images from severely degraded measurements has become an important research problem, with the goal of preserving diagnostic quality while minimizing patient exposure~\cite{yu2025robust,jang2023spachtrans,yang2023drmc}.

Deep learning has advanced low-dose PET denoising, progressing through several architectural paradigms. Early CNN–based models~\cite{wang2021artificial,chen2017RED-CNN,liang2020edcnn,cciccek20163d-unet} achieved effective noise suppression but often produced over-smoothed results with loss of fine anatomical detail. 
GAN-based approaches~\cite{luo2022argan,wang20183D-cGAN,zhou2021mdpet} improved perceptual realism through adversarial training, yet frequently suffered from training instability and hallucination risk. 
More recently, transformer-based architectures~\cite{zamir2022restormer,jang2023spachtrans} demonstrated stronger global context modeling, but their high memory and computational cost limit practical deployment for large 3D whole-body PET volumes. 
Denoising Diffusion Probabilistic Models (DDPM)~\cite{ho2020ddpm} have therefore emerged as a promising alternative due to their stable optimization and strong generative fidelity. 
Recent diffusion-based PET denoising frameworks~\cite{gong2024pet,lyu2025widpet,yu2025robust,jing2026wcc} now achieve state-of-the-art performance across multiple benchmarks.
\begin{figure*}[t!]
\centering
\includegraphics[width=\linewidth]{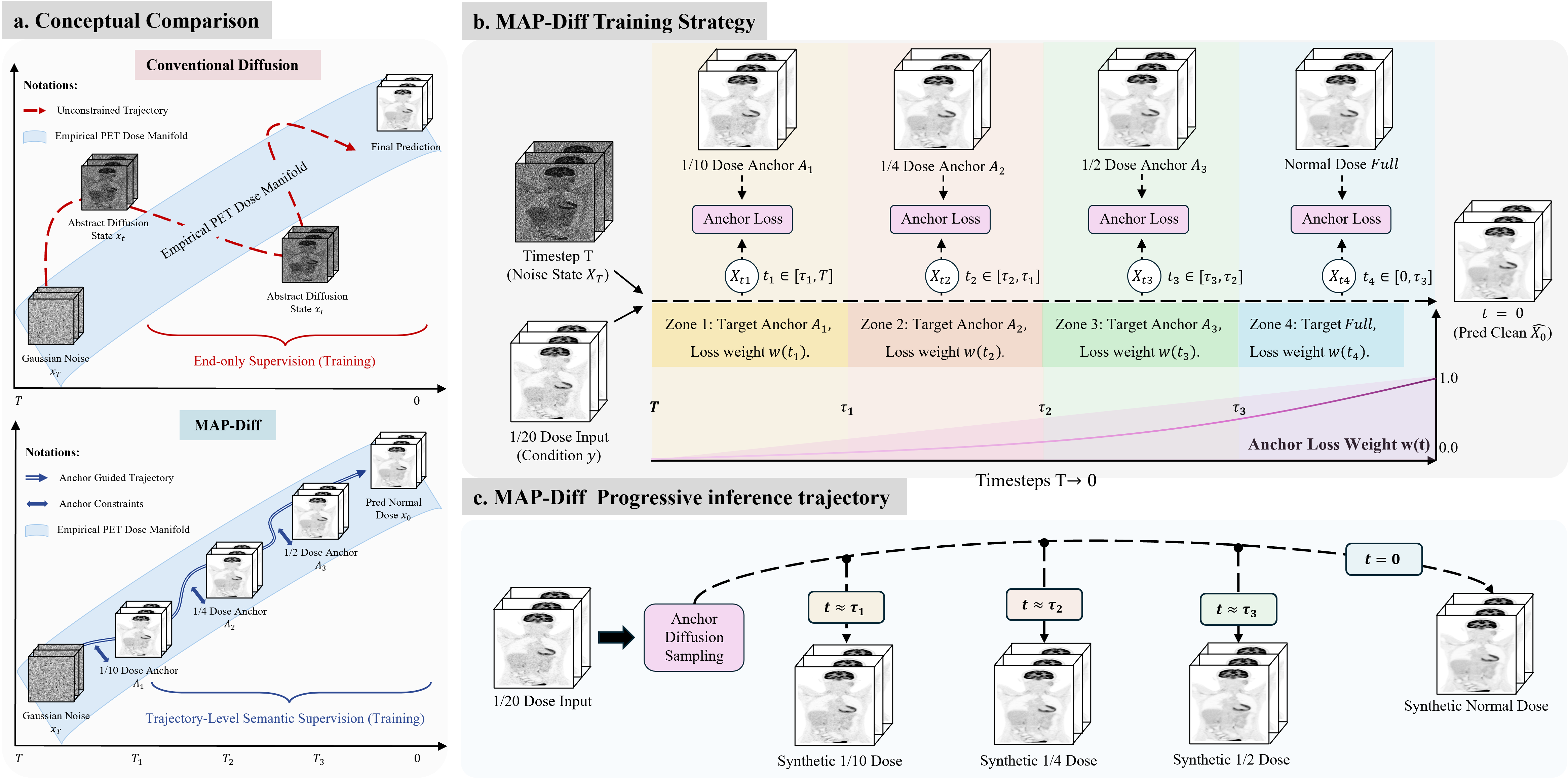}
\caption{
Overview of MAP-Diff.
(a) Conceptual comparison between standard diffusion training with end-point supervision and the proposed
multi-anchor trajectory-level supervision.
(b) MAP-Diff training strategy with timestep zoning and anchor-weighted loss.
(c) Progressive inference trajectory enabling dose-consistent intermediate outputs.
}
\label{fig:method}
\end{figure*}

Despite this progress, diffusion-based PET denoising still exhibits two key mismatches between training objectives and the underlying data structure. \textbf{First}, standard diffusion models are primarily optimized for final reconstruction accuracy at $t=0$, while the intermediate reverse trajectory remains unconstrained. As illustrated in Fig.~\ref{fig:method}(a), the denoising path is learned implicitly through noise prediction and is not explicitly aligned with clinically meaningful dose levels, leading to dose-inconsistent intermediate states. 
As a result, intermediate states may satisfy the diffusion objective but not semantically dose-consistent.
\textbf{Second}, PET image formation is inherently progressive: image quality improves gradually with increasing count statistics, and intermediate-dose images follow structured, non-arbitrary appearance patterns governed by acquisition physics and tracer kinetics~\cite{berger2003positron,czernin2002positron}. However, existing diffusion approaches do not explicitly encode this progressive dose structure in the reverse process. Even when multi-dose data are available, supervision is typically applied only at the final target, leaving the reverse trajectory physically unregularized and 
failing to exploit structured intermediate-dose information. This creates a structural mismatch: the learned generative trajectory is unconstrained with respect to the monotonic and physically grounded dose progression inherent to PET acquisition.
 
To address these two mismatches, we propose MAP-Diff (Fig.~\ref{fig:method}(b)), a multi-anchor guided diffusion framework for progressive 3D whole-body low-dose PET denoising. MAP-Diff introduces clinically observed intermediate-dose images as \emph{trajectory anchors} and applies timestep-dependent supervision along the reverse diffusion process. 
By extending supervision from the final output to the full denoising trajectory, MAP-Diff resolves the objective mismatch. Simultaneously, by enforcing stage-wise alignment with physically grounded dose levels, it regularizes the generative path to reflect the progressive nature of PET acquisition.

As a result, the reverse diffusion trajectory becomes both semantically structured and dose-consistent across clinically meaningful stages, while still requiring only ultra-low-dose input at inference.
Our main contributions are threefold:
\begin{itemize}
  \item We propose MAP-Diff, a multi-anchor guided diffusion framework that constrains the reverse denoising trajectory using clinically observed intermediate-dose anchors, rather than supervising only the final full-dose target.

  \item We introduce timestep-dependent trajectory supervision  that aligns diffusion corruption with real PET dose degradation, 
  enforcing trajectory-level consistency and enabling dose-consistent intermediate reconstructions.

  \item Comprehensive evaluation on ultra-low-dose 3D whole-body PET across internal and cross-scanner datasets, showing consistent gains in structural fidelity and quantitative accuracy over state-of-the-art baselines.

\end{itemize}

\section{Method}

\subsection{Problem Setup and Notation}\label{subsec:problem_setup}
Let $\Omega \subset \mathbb{Z}^3$ denote a 3D voxel grid of size $H \times W \times D$. 
For each subject, we observe paired multi-dose PET volumes
$\{x^{1/20}, x^{1/10}, x^{1/4}, x^{1/2}, x^{\text{full}}\}$. 
We define the ultra-low-dose image as the conditional input $y = x^{1/20}$ and denote the full-dose reference as $x_0 := x^{\text{full}}$. To model the progressive nature of PET dose formation, we introduce a set of clinically observed \textbf{dose anchors}
$\mathcal{A} = \{a_1, a_2, a_3, a_4\}$,
where $a_1,a_2,a_3$ correspond to the $1/10$, $1/4$, and $1/2$ dose volumes, and $a_4 = x_0$.
Our goal is to learn a conditional generative model that restores full-dose PET from ultra-low-dose input while enforcing a dose-consistent denoising trajectory. Only $y$ is required at inference time.

\subsection{Conditional Diffusion Backbone}\label{subsec:conditional_diffusion}
We employ a conditional denoising diffusion probabilistic model (DDPM)~\cite{ho2020ddpm}. 
Given a variance schedule $\{\alpha_t\}_{t=1}^T$, the forward process admits the closed-form marginal
\begin{equation}
x_t = \sqrt{\bar{\alpha}_t}\,x_0 + \sqrt{1-\bar{\alpha}_t}\,\epsilon, 
\quad \epsilon \sim \mathcal{N}(0,\mathbf{I}),
\end{equation}
where $\bar{\alpha}_t = \prod_{s=1}^{t} \alpha_s$. Although PET noise follows Poisson photon-counting statistics~\cite{berger2003positron,czernin2002positron}, we adopt Gaussian diffusion corruption as a tractable surrogate in image space; boundary calibration (Sec.~\ref{subsec:bounday_calibration}) aligns statistical degradation levels rather than assuming physical noise equivalence.

A conditional denoiser $\epsilon_\theta(x_t,t\mid y)$ predicts the injected noise given the noisy volume, timestep, and ultra-low-dose condition. 
A per-timestep clean estimate is obtained as
\begin{equation}
\hat{x}_0(x_t,t\mid y) =
\frac{1}{\sqrt{\bar{\alpha}_t}}
\left(x_t - \sqrt{1-\bar{\alpha}_t}\,\epsilon_\theta(x_t,t\mid y)\right).
\end{equation}
Unlike standard diffusion training that uses $\hat{x}_0$ only implicitly, MAP-Diff uses this estimate for explicit trajectory-level supervision.

\subsection{Multi-Anchor Trajectory Supervision}\label{subsec:multi_anchor}
To resolve the objective and structure mismatches discussed in Sec.~\ref{sec:intro}, MAP-Diff (Fig.~\ref{fig:method} (b)) introduces trajectory-level supervision using clinically observed dose anchors. Instead of supervising only the final reconstruction target, we constrain the reverse diffusion trajectory itself to follow dose-aligned stages. We partition the timestep axis into four supervision zones with boundaries
$0 \le \tau_3 < \tau_2 < \tau_1 < T$ and define intervals
$
\mathcal{I}_1 = [\tau_1,T], \;
\mathcal{I}_2 = [\tau_2,\tau_1), \;
\mathcal{I}_3 = [\tau_3,\tau_2), \;
\mathcal{I}_4 = [0,\tau_3).
$
Each timestep is assigned an anchor target through a piecewise mapping: $a(t) = \sum_{j=1}^{4} \mathbf{1}[t\in\mathcal{I}_j]\,a_j$.

This zoning strategy defines a dose-aligned semantic curriculum along the reverse process: high-noise timesteps are guided toward lower-dose anchors, while later timesteps are constrained toward higher-dose targets. As a result, intermediate states are explicitly regularized to be dose-consistent rather than unconstrained latent denoising states.

\subsubsection{Boundary Calibration}\label{subsec:bounday_calibration}
The timestep boundaries $(\tau_1,\tau_2,\tau_3)$ are calibrated by matching diffusion corruption levels to clinically observed dose degradation. 
Within a body mask $M$, we compute degradation signatures relative to full dose using both intensity and structural criteria:
\begin{align}
d^{\text{clin}}_j &=
\Big(
\mathrm{NMAE}\big(
M \odot a_j,\;
M \odot x_0
\big), \nonumber\\
&\quad
1 - \mathrm{SSIM}\big(
M \odot a_j,\;
M \odot x_0
\big)
\Big), \\
d^{\text{sim}}(t) &=
\Big(
\mathrm{NMAE}\big(
M \odot x_t,\;
M \odot x_0
\big), \nonumber\\
&\quad
1 - \mathrm{SSIM}\big(
M \odot x_t,\;
M \odot x_0
\big)
\Big).
\end{align}
Matched timesteps are selected by
$t_j^\star = \arg\min_t \|d^{\text{sim}}(t)-d^{\text{clin}}_j\|_2$,
and final boundaries are obtained from dataset-level averages.

\subsection{Training Objective and Inference}\label{subsec:inference}
Training combines standard diffusion noise prediction with anchor-guided trajectory supervision. 
Sampling $t \sim \mathrm{Unif}\{1,\dots,T\}$, the standard objective is
\begin{equation}
\mathcal{L}_{\text{noise}} =
\mathbb{E}_{t,\epsilon}
\bigl[\|\epsilon - \epsilon_\theta(x_t,t\mid y)\|_2^2\bigr],
\end{equation}
which ensures correct diffusion modeling but supervises only the final target.

To extend supervision to the trajectory and enforce dose consistency, we add an anchor reconstruction loss on the per-timestep clean estimate:
\begin{equation}
\mathcal{L}_{\text{anch}} =
\mathbb{E}_{t}
\bigl[w(t)\|\hat{x}_0(x_t,t\mid y)-a(t)\|_2^2\bigr].
\end{equation}

The timestep weight $w(t)=(1-t/T)^p$ emphasizes later, lower-noise stages where predictions are more reliable, avoiding over-constraining early noisy states while preserving progressive alignment. 

We adopt hard timestep zoning and a polynomial weighting schedule
as a simple and stable mechanism for stage-aware supervision.
In practice, soft assignments or learned weighting functions could
also be employed; however, the piecewise formulation provides
clear dose-stage separation and was found empirically sufficient
without introducing additional optimization complexity.The final loss is:
\begin{equation}
\mathcal{L} =
\mathcal{L}_{\text{noise}} + \lambda \mathcal{L}_{\text{anch}}.
\end{equation}

The addition of the anchor term modifies the standard DDPM objective.
While $\mathcal{L}_{\text{noise}}$ preserves the score-matching formulation,
$\mathcal{L}_{\text{anch}}$ acts as a trajectory-level regularizer.
Thus, MAP-Diff can be viewed as a regularised conditional score model
rather than a strictly likelihood-consistent diffusion model,
prioritizing dose-consistent trajectory shaping over exact likelihood optimization.

At inference (Fig.~\ref{fig:method} (c)), given only input $y$, we run the reverse diffusion process to obtain the final full-dose estimate $\hat{x}_0$. Because supervision is applied along the trajectory, intermediate-dose synthesis can be exported by recording $\hat{x}_0(x_t,t\mid y)$ at $t=\tau_1,\tau_2,\tau_3$, yielding synthetic volumes aligned with intermediate dose levels.

\section{Experiments and Results}\label{sec:experiments}
\subsection{Experiments}
\paragraph{Dataset.}
We use the Ultra-Low-Dose PET Challenge dataset~\cite{xue2025udpet}, which provides multi-dose whole-body PET scans from two vendors: University Hospital of Bern (Siemens Biograph Vision Quadra) and Shanghai Ruijin Hospital (United Imaging uEXPLORER). For in-distribution experiments, we select 377 $^{18}$F-FDG scans from the Siemens system with paired dose levels (1/20, 1/10, 1/4, 1/2, and normal). Subjects are split into 297/20/60 for training/validation/testing. For cross-scanner evaluation, we use 100 uEXPLORER scans for external testing only. All volumes are converted to SUV, center-cropped to $192\times288\times520$, and divided into overlapping $96\times96\times96$ patches for fully 3D training~\cite{yu2025robust}.
\begin{table*}[t!]
\centering
\caption{Quantitative comparison results. Statistical significance is assessed using a two-sided Wilcoxon signed-rank test with Holm correction. * and ** denote statistically significant differences compared with MAP-Diff at $p<0.05$ and $p<0.01$, respectively.}
\setlength{\tabcolsep}{4pt}

\resizebox{\linewidth}{!}{
\begin{tabular}{lccc|ccc}
\toprule
\textbf{Method} 
& \multicolumn{3}{c|}{\textbf{Internal (University Hospital of Bern)}} 
& \multicolumn{3}{c}{\textbf{External (Shanghai Ruijin Hospital)}} \\
\cmidrule(lr){2-4} \cmidrule(lr){5-7}
& \textbf{PSNR$\uparrow$} & \textbf{SSIM$\uparrow$} & \textbf{NMAE$\downarrow$}
& \textbf{PSNR$\uparrow$} & \textbf{SSIM$\uparrow$} & \textbf{NMAE$\downarrow$} \\
\midrule

Low-dose Input
& 38.545$\pm$1.798** & 0.948$\pm$0.018** & 0.199$\pm$0.036**
& 30.402$\pm$3.475** & 0.896$\pm$0.033** & 0.257$\pm$0.059** \\

\midrule

3D cGAN~\cite{wang20183D-cGAN}
& 40.721$\pm$1.293** & 0.977$\pm$0.006** & 0.127$\pm$0.015**
& 33.062$\pm$3.624** & 0.927$\pm$0.053** & 0.174$\pm$0.031** \\

3D UNet~\cite{cciccek20163d-unet}
& 41.810$\pm$1.377** & 0.972$\pm$0.007** & 0.120$\pm$0.015**
& 33.654$\pm$3.810** & 0.904$\pm$0.074** & 0.164$\pm$0.031** \\

RED-CNN~\cite{chen2017RED-CNN}
& 42.185$\pm$1.427** & 0.981$\pm$0.005** & 0.119$\pm$0.016**
& 33.989$\pm$3.725** & 0.945$\pm$0.025** & 0.161$\pm$0.030** \\

EDCNN~\cite{liang2020edcnn}
& 42.355$\pm$1.579** & 0.983$\pm$0.006** & 0.116$\pm$0.017**
& 33.841$\pm$3.615** & 0.942$\pm$0.023** & 0.164$\pm$0.024** \\

LIT-Former~\cite{chen2024lit}
& 42.036$\pm$1.537** & 0.981$\pm$0.006** & 0.121$\pm$0.017**
& 33.471$\pm$3.723** & 0.931$\pm$0.037** & 0.176$\pm$0.042** \\

Restormer~\cite{zamir2022restormer}
& 42.198$\pm$1.466** & 0.979$\pm$0.007** & 0.124$\pm$0.020**
& 33.657$\pm$3.811** & 0.939$\pm$0.027** & 0.171$\pm$0.036** \\

Spach Former~\cite{jang2023spachtrans}
& 42.290$\pm$1.515** & 0.982$\pm$0.005** & 0.118$\pm$0.015**
& 34.109$\pm$3.882** & 0.945$\pm$0.030** & 0.153$\pm$0.032** \\

DRMC~\cite{yang2023drmc}
& 42.378$\pm$1.619** & 0.982$\pm$0.006** & 0.115$\pm$0.018**
& 34.006$\pm$3.832** & 0.942$\pm$0.027** & 0.163$\pm$0.034** \\

3D DDPM~\cite{yu2025robust}
& 42.483$\pm$1.593** & 0.983$\pm$0.005** & 0.115$\pm$0.016**
& 34.117$\pm$3.485** & 0.947$\pm$0.021** & 0.151$\pm$0.023** \\

\midrule
\textbf{MAP-Diff}
& \textbf{43.710$\pm$1.575}
& \textbf{0.986$\pm$0.004}
& \textbf{0.103$\pm$0.014}
& \textbf{34.419$\pm$3.611}
& \textbf{0.952$\pm$0.023}
& \textbf{0.141$\pm$0.025} \\

\bottomrule
\end{tabular}}
\label{tab:quant_results}
\end{table*}
\begin{figure*}[t!]
\centering
\includegraphics[width=\linewidth]{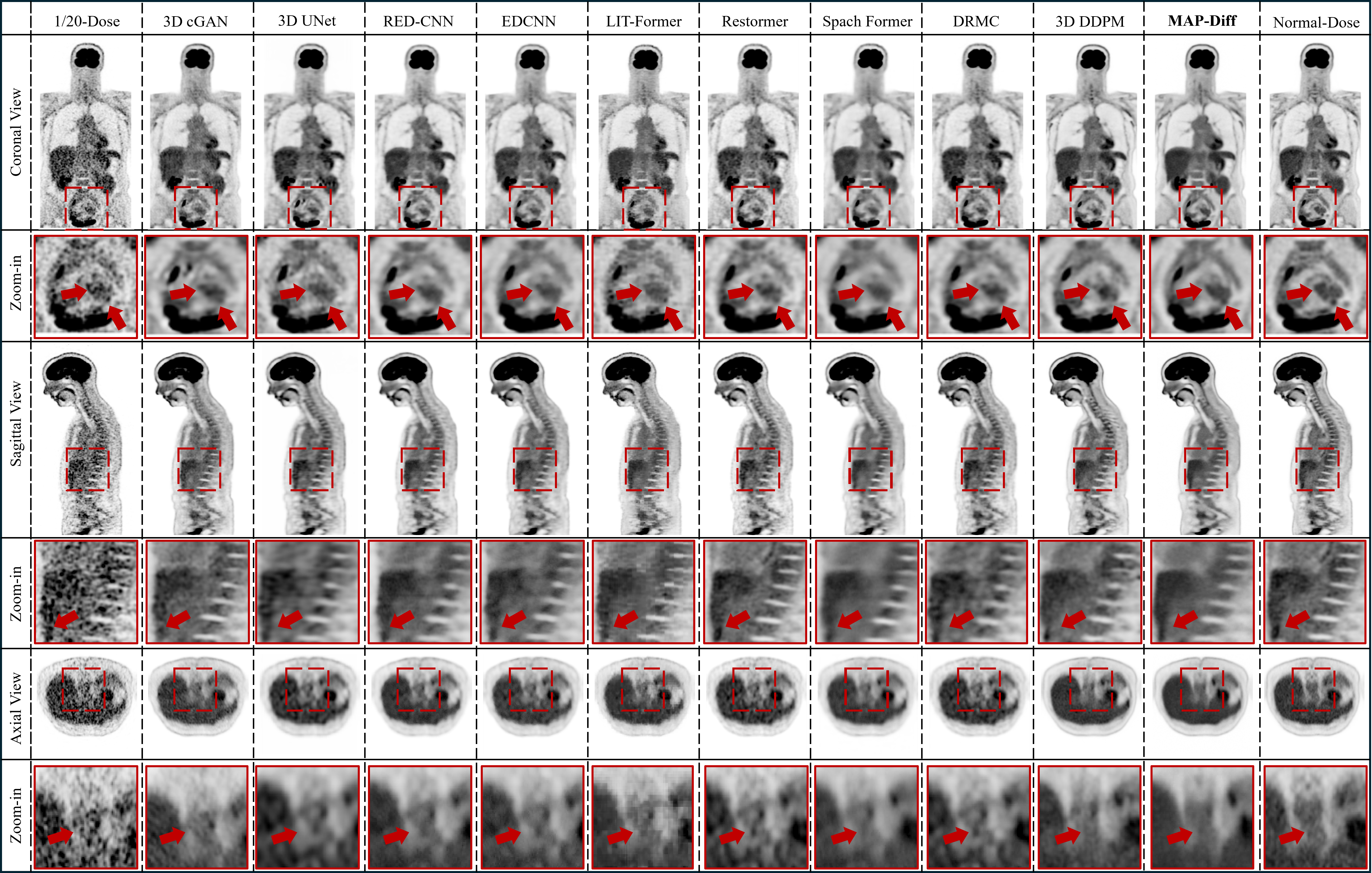}
\caption{Visual comparison  of PET denoising across methods.}

\label{fig:comp}
\end{figure*}

\paragraph{Evaluation Metrics.} We use PSNR~\cite{hore2010image} and SSIM~\cite{psnrssim} with normal-dose PET as the golden-rule. PSNR measures voxel-wise fidelity, while SSIM assesses structural and perceptual similarity. We additionally report NMAE~\cite{psnrssim} to quantify intensity deviations, which is particularly important in PET since all volumes are expressed in SUV units and require accurate quantitative recovery. 

\paragraph{Experiment Details.}
All experiments are implemented in PyTorch and run on NVIDIA A100 GPUs. Diffusion models use $T{=}1000$ timesteps with a linear noise schedule ($\beta_t$ from $1{\times}10^{-4}$ to $2{\times}10^{-2}$) and are optimized with Adam~\cite{zhang2018improved} at learning rate $1{\times}10^{-4}$. All methods are evaluated in fully 3D form for volumetric PET. Baselines originally proposed in 2D~\cite{liang2020edcnn,chen2017RED-CNN,chen2024lit,zamir2022restormer} are converted to 3D while preserving their original architectures and training settings.

\subsection{Results}
\paragraph{Overall quantitative and qualitative performance.}
Table~\ref{tab:quant_results} reports quantitative comparisons on the internal (Siemens Quadra) and external (uEXPLORER) cohorts. MAP-Diff achieves the best performance across all metrics on both datasets, with statistically significant improvements over all competing methods ($p<0.01$).
On the internal cohort, MAP-Diff increases PSNR to 43.71$\pm$1.58\,dB, outperforming the strongest diffusion baseline (3D DDPM~\cite{yu2025robust}) by +1.23\,dB. NMAE is reduced from 0.115 to 0.103, while SSIM improves to 0.986, depicting enhanced structural fidelity and quantitative accuracy.
On the external cohort, MAP-Diff achieves 34.42$\pm$3.61\,dB PSNR and reduces NMAE to 0.141, again surpassing 3D DDPM (34.12\,dB, 0.151 NMAE) and all CNN-~\cite{cciccek20163d-unet,chen2017RED-CNN,liang2020edcnn,yang2023drmc}, GAN-~\cite{wang20183D-cGAN}, and Transformer-~\cite{chen2024lit,zamir2022restormer,jang2023spachtrans} based baselines.
Qualitatively (Fig.~\ref{fig:comp}), MAP-Diff better preserves lesion contrast and anatomical boundaries while effectively suppressing noise without introducing over-smoothing artifacts. 

\begin{figure*}[t!]
\centering
\includegraphics[width=0.95\linewidth]{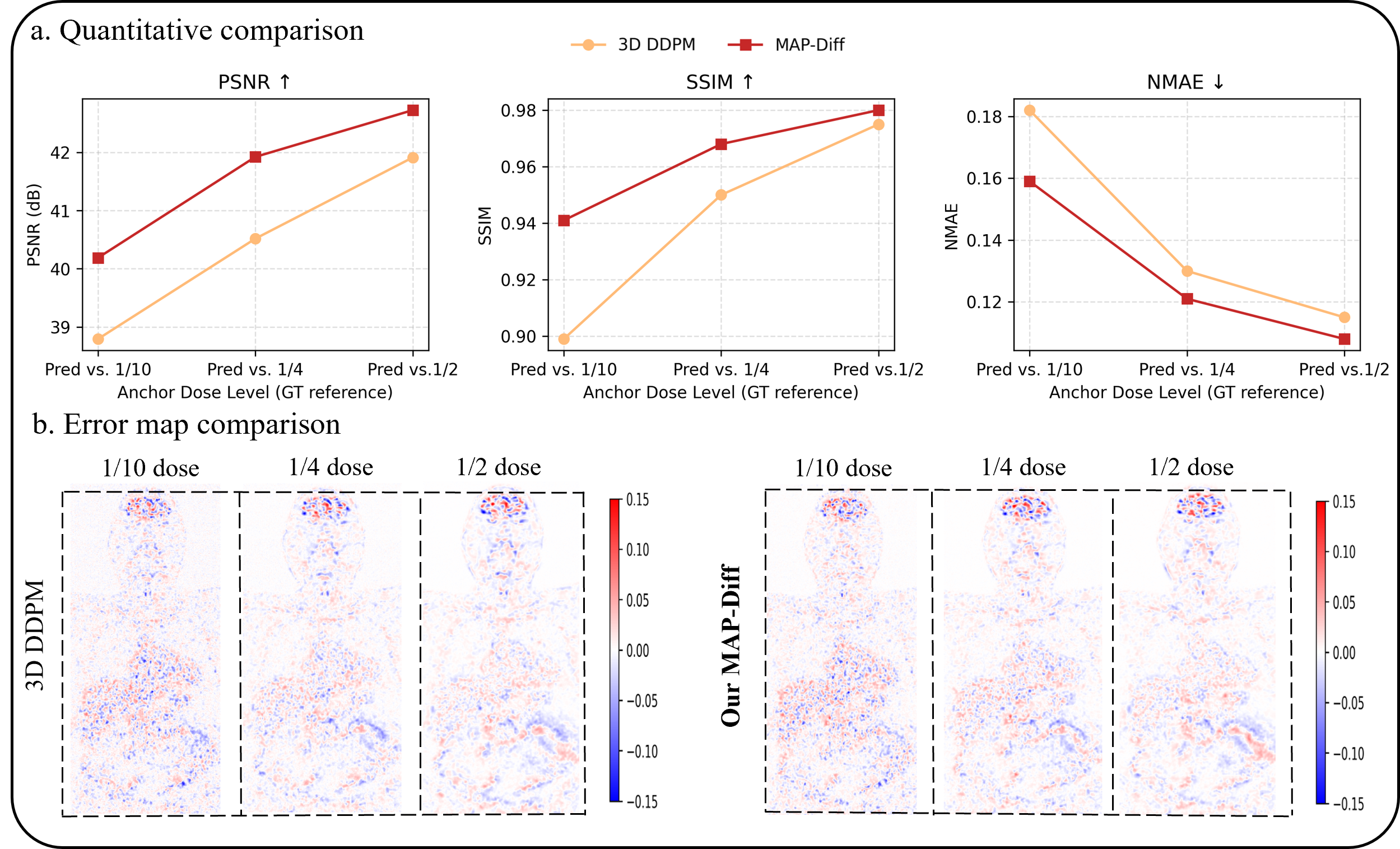}
\caption{
Intermediate-dose reconstruction accuracy at anchor timesteps.
(a) Quantitative comparison against paired real-dose references at
the 1/10, 1/4, and 1/2 dose levels.
(b) Coronal-view signed error maps against paired real-dose references.
}
\label{fig:inter}
\end{figure*}

\begin{table*}[h!]
\centering
\caption{Ablation study of anchor supervision design in MAP-Diff, evaluating the number of anchors and timestep-weighted loss. Best results in bold.}
\label{tab:anchor_ablation}
\setlength{\tabcolsep}{3pt}
\renewcommand{\arraystretch}{0.95}
\footnotesize
\resizebox{\linewidth}{!}{
\begin{tabular}{l c l c c c c @{\hspace{4pt}}|@{\hspace{4pt}} l c l c c c c}
\toprule
\textbf{Set} & \textbf{\#A} & \textbf{Anchors} & \textbf{W} & \textbf{PSNR} & \textbf{SSIM} & \textbf{NMAE}
& \textbf{Set} & \textbf{\#A} & \textbf{Anchors} & \textbf{W} & \textbf{PSNR} & \textbf{SSIM} & \textbf{NMAE} \\
\midrule
\multicolumn{14}{c}{Baseline 3D DDPM~\cite{yu2025robust} (No Anchors): PSNR: 42.483, SSIM: 0.983, NMAE: 0.115} \\
\midrule
\textbf{S1} & 3 & \{1/10,1/4,1/2\} & Poly & \textbf{43.710} & \textbf{0.986} & \textbf{0.103} & S2 & 2 & \{1/10,1/4\} & Poly & 43.672 & 0.984 & 0.104 \\
S3 & 2 & \{1/10,1/2\} & Poly & 43.521 & 0.983 & 0.106 & S4 & 2 & \{1/4,1/2\} & Poly & 43.414 & 0.982 & 0.107 \\
S5 & 1 & \{1/10\} & Poly & 43.503 & 0.983 & 0.106 & S6 & 1 & \{1/4\} & Poly & 43.267 & 0.982 & 0.109 \\
S7 & 1 & \{1/2\} & Poly & 42.790 & 0.981 & 0.111 & S8 & 3 & \{1/10,1/4,1/2\} & Const & 41.856 & 0.905 & 0.133 \\
\bottomrule
\end{tabular}}
\end{table*}

\paragraph{Intermediate-dose reconstruction behavior.}
To validate the progressive reconstruction property described in Sec.~\ref{subsec:inference}, we compare intermediate outputs extracted at anchor timesteps with real paired intermediate-dose references. As shown in Fig.~\ref{fig:inter} (a), MAP-Diff consistently outperforms 3D DDPM at the $1/10$, $1/4$, and $1/2$ dose levels across PSNR, SSIM, and NMAE. The largest margin occurs at the earliest stage ($1/10$ dose), where noise levels are highest and trajectory guidance is most critical. The error maps in Fig.~\ref{fig:inter}(b) further show that MAP-Diff produces more spatially homogeneous residuals with reduced structured errors compared with 3D DDPM across all stages. This behaviour indicates that anchor-based supervision provides effective stage-local constraints, steering the reverse process toward dose-consistent intermediate states. As a result, MAP-Diff improves not only the final reconstruction but also the quantitative accuracy of intermediate outputs along the trajectory.

\paragraph{Ablation on anchor supervision design.}
Table~\ref{tab:anchor_ablation} presents ablations of the anchor supervision design, varying both the number of anchors and the timestep weighting strategy. Using all three anchors with timestep-weighted loss (S1) yields the best overall performance across all metrics. Reducing the number of anchors leads to gradual degradation, with configurations that include the earliest anchor (1/10 dose) consistently outperforming those that use only later anchors, indicating that early-stage guidance contributes most to trajectory shaping. Replacing timestep-dependent weighting with constant weighting (S8) significantly degrades performance, suggesting that stage-aware loss scaling is important to stabilize training and avoid over-constraining high-noise timesteps. Together, these results confirm that both multi-anchor coverage and timestep-adaptive weighting are key to effective trajectory supervision.










\section{Conclusion}\label{sec:conclusion}
We introduced MAP-Diff, a multi-anchor guided diffusion framework for progressive 3D whole-body low-dose PET denoising. By introducing clinically observed intermediate-dose anchors and timestep-dependent trajectory supervision, MAP-Diff constrains the reverse diffusion process to follow a dose-aligned and stage-consistent restoration path, addressing the mismatch between endpoint-only objectives and progressive PET dose structure. The anchor-based trajectory regularization enables stage-wise semantic consistency while requiring only ultra-low-dose input at inference. Experiments on both internal and cross-scanner datasets demonstrate consistent improvements over strong baselines in visual quality and quantitative accuracy. Future work will explore adaptive anchor selection and extension to other progressive medical image reconstruction settings.

\section*{Acknowledgements}
Javier Montoya is supported by the Swiss National Science Foundation (SNSF) under grant number 20HW-1 220785. Guang Yang was supported in part by the ERC IMI (101005122), the H2020 (952172), the MRC (MC/PC/21013), the Royal Society (IEC/NSFC/211235), the NVIDIA Academic Hardware Grant Program, the SABER project supported by Boehringer Ingelheim Ltd, NIHR Imperial Biomedical Research Centre (RDA01), The Wellcome Leap Dynamic resilience program (co-funded by Temasek Trust)., UKRI guarantee funding for Horizon Europe MSCA Postdoctoral Fellowships (EP/Z002206/1), UKRI MRC Research Grant, TFS Research Grants (MR/U506710/1), Swiss National Science Foundation (Grant No. 220785), and the UKRI Future Leaders Fellowship (MR/V023799/1, UKRI2738). Peiyuan Jing is supported by the Swiss National Science Foundation (SNSF) under grant number 20HW-1 220785.
\nocite{langley00}

\bibliography{example_paper}

@article{berger2003positron,
  title={Positron emission tomography},
  author={Berger, Abi},
  journal={BMJ: British Medical Journal},
  volume={326},
  number={7404},
  pages={1449},
  year={2003}
}

@inproceedings{hore2010image,
  title={Image quality metrics: PSNR vs. SSIM},
  author={Hore, Alain and Ziou, Djemel},
  booktitle={2010 20th international conference on pattern recognition},
  pages={2366--2369},
  year={2010},
  organization={IEEE}
}

@inproceedings{zhang2018improved,
  title={Improved adam optimizer for deep neural networks},
  author={Zhang, Zijun},
  booktitle={2018 IEEE/ACM 26th international symposium on quality of service (IWQoS)},
  pages={1--2},
  year={2018},
  organization={Ieee}
}

@article{czernin2002positron,
  title={Positron emission tomography scanning: current and future applications},
  author={Czernin, Johannes and Phelps, Michael E},
  journal={Annual review of medicine},
  volume={53},
  number={1},
  pages={89--112},
  year={2002},
  publisher={Annual Reviews 4139 El Camino Way, PO Box 10139, Palo Alto, CA 94303-0139, USA}
}

@article{fletcher2008recommendations,
  title={Recommendations on the use of 18F-FDG PET in oncology},
  author={Fletcher, James W and Djulbegovic, Benjamin and Soares, Heloisa P and Siegel, Barry A and Lowe, Val J and Lyman, Gary H and Coleman, R Edward and Wahl, Richard and Paschold, John Christopher and Avril, Norbert and others},
  journal={Journal of Nuclear Medicine},
  volume={49},
  number={3},
  pages={480--508},
  year={2008},
  publisher={Society of Nuclear Medicine}
}

@article{wang2021artificial,
  title={Artificial intelligence enables whole-body positron emission tomography scans with minimal radiation exposure},
  author={Wang, Yan-Ran and Baratto, Lucia and Hawk, K Elizabeth and Theruvath, Ashok J and Pribnow, Allison and Thakor, Avnesh S and Gatidis, Sergios and Lu, Rong and Gummidipundi, Santosh E and Garcia-Diaz, Jordi and others},
  journal={European journal of nuclear medicine and molecular imaging},
  volume={48},
  number={9},
  pages={2771--2781},
  year={2021},
  publisher={Springer}
}

@inproceedings{liang2020edcnn,
  title={Edcnn: Edge enhancement-based densely connected network with compound loss for low-dose ct denoising},
  author={Liang, Tengfei and Jin, Yi and Li, Yidong and Wang, Tao},
  booktitle={2020 15th IEEE International conference on signal processing (ICSP)},
  volume={1},
  pages={193--198},
  year={2020},
  organization={IEEE}
}

@article{chen2017RED-CNN,
  title={Low-dose CT with a residual encoder-decoder convolutional neural network},
  author={Chen, Hu and Zhang, Yi and Kalra, Mannudeep K and Lin, Feng and Chen, Yang and Liao, Peixi and Zhou, Jiliu and Wang, Ge},
  journal={IEEE transactions on medical imaging},
  volume={36},
  number={12},
  pages={2524--2535},
  year={2017},
  publisher={IEEE}
}

@article{wang20183D-cGAN,
  title={3D conditional generative adversarial networks for high-quality PET image estimation at low dose},
  author={Wang, Yan and Yu, Biting and Wang, Lei and Zu, Chen and Lalush, David S and Lin, Weili and Wu, Xi and Zhou, Jiliu and Shen, Dinggang and Zhou, Luping},
  journal={Neuroimage},
  volume={174},
  pages={550--562},
  year={2018},
  publisher={Elsevier}
}

@inproceedings{yang2023drmc,
  title={Drmc: A generalist model with dynamic routing for multi-center pet image synthesis},
  author={Yang, Zhiwen and Zhou, Yang and Zhang, Hui and Wei, Bingzheng and Fan, Yubo and Xu, Yan},
  booktitle={International Conference on Medical Image Computing and Computer-Assisted Intervention},
  pages={36--46},
  year={2023},
  organization={Springer}
}

@article{luo2022argan,
  title={Adaptive rectification based adversarial network with spectrum constraint for high-quality PET image synthesis},
  author={Luo, Yanmei and Zhou, Luping and Zhan, Bo and Fei, Yuchen and Zhou, Jiliu and Wang, Yan and Shen, Dinggang},
  journal={Medical image analysis},
  volume={77},
  pages={102335},
  year={2022},
  publisher={Elsevier}
}

@inproceedings{cciccek20163d-unet,
  title={3D U-Net: learning dense volumetric segmentation from sparse annotation},
  author={{\c{C}}i{\c{c}}ek, {\"O}zg{\"u}n and Abdulkadir, Ahmed and Lienkamp, Soeren S and Brox, Thomas and Ronneberger, Olaf},
  booktitle={International conference on medical image computing and computer-assisted intervention},
  pages={424--432},
  year={2016},
  organization={Springer}
}

@article{jang2023spachtrans,
  title={Spach Transformer: Spatial and channel-wise transformer based on local and global self-attentions for PET image denoising},
  author={Jang, Se-In and Pan, Tinsu and Li, Ye and Heidari, Pedram and Chen, Junyu and Li, Quanzheng and Gong, Kuang},
  journal={IEEE transactions on medical imaging},
  volume={43},
  number={6},
  pages={2036--2049},
  year={2023},
  publisher={IEEE}
}

@inproceedings{zamir2022restormer,
  title={Restormer: Efficient transformer for high-resolution image restoration},
  author={Zamir, Syed Waqas and Arora, Aditya and Khan, Salman and Hayat, Munawar and Khan, Fahad Shahbaz and Yang, Ming-Hsuan},
  booktitle={Proceedings of the IEEE/CVF conference on computer vision and pattern recognition},
  pages={5728--5739},
  year={2022}
}

@article{yu2025robust,
  title={Robust whole-body PET image denoising using 3D diffusion models: evaluation across various scanners, tracers, and dose levels},
  author={Yu, Boxiao and Ozdemir, Savas and Dong, Yafei and Shao, Wei and Pan, Tinsu and Shi, Kuangyu and Gong, Kuang},
  journal={European Journal of Nuclear Medicine and Molecular Imaging},
  volume={52},
  number={7},
  pages={2549--2562},
  year={2025},
  publisher={Springer}
}

@inproceedings{lyu2025widpet,
  title={WiD-PET: PET Image Reconstruction from Low-Dose Data Using a Wavelet-Informed Diffusion Model with Fast Inference},
  author={Lyu, Qingcheng and Chen, Tong and Wang, Yiran and Guo, Erjian and Zhou, Luping},
  booktitle={International Conference on Medical Image Computing and Computer-Assisted Intervention},
  pages={684--693},
  year={2025},
  organization={Springer}
}

@article{jing2026wcc,
  title={3D Wavelet-Based Structural Priors for Controlled Diffusion in Whole-Body Low-Dose PET Denoising},
  author={Jing, Peiyuan and Tang, Yue and Cheng, Chun-Wun and Zhang, Zhenxuan and Yang, Liutao and Lima, Thiago V and Strobel, Klaus and Leimgruber, Antoine and Aviles-Rivero, Angelica and Yang, Guang and others},
  journal={arXiv preprint arXiv:2601.07093},
  year={2026}
}

@article{zhou2021mdpet,
  title={MDPET: a unified motion correction and denoising adversarial network for low-dose gated PET},
  author={Zhou, Bo and Tsai, Yu-Jung and Chen, Xiongchao and Duncan, James S and Liu, Chi},
  journal={IEEE transactions on medical imaging},
  volume={40},
  number={11},
  pages={3154--3164},
  year={2021},
  publisher={IEEE}
}

@article{ho2020ddpm,
  title={Denoising diffusion probabilistic models},
  author={Ho, Jonathan and Jain, Ajay and Abbeel, Pieter},
  journal={Advances in neural information processing systems},
  volume={33},
  pages={6840--6851},
  year={2020}
}

@article{gong2024pet,
  title={PET image denoising based on denoising diffusion probabilistic model},
  author={Gong, Kuang and Johnson, Keith and El Fakhri, Georges and Li, Quanzheng and Pan, Tinsu},
  journal={European Journal of Nuclear Medicine and Molecular Imaging},
  volume={51},
  number={2},
  pages={358--368},
  year={2024},
  publisher={Springer}
}

@inproceedings{xue2025udpet,
  title={UDPET: Ultra-low Dose PET Imaging Challenge Dataset},
  author={Xue, Song and Wang, Hanzhong and Chen, Yizhou and Liu, Fanxuan and Zhu, Hong and Viscione, Marco and Guo, Rui and Rominger, Axel and Li, Biao and Shi, Kuangyu},
  booktitle={International Conference on Medical Image Computing and Computer-Assisted Intervention},
  pages={616--623},
  year={2025},
  organization={Springer}
}

@article{chen2024lit,
  title={LIT-Former: Linking in-plane and through-plane transformers for simultaneous CT image denoising and deblurring},
  author={Chen, Zhihao and Niu, Chuang and Gao, Qi and Wang, Ge and Shan, Hongming},
  journal={IEEE Transactions on Medical Imaging},
  volume={43},
  number={5},
  pages={1880--1894},
  year={2024},
  publisher={IEEE}
}

@article{psnrssim,
  title={Image quality assessment: from error visibility to structural similarity},
  author={Wang, Zhou and Bovik, Alan C and Sheikh, Hamid R and Simoncelli, Eero P},
  journal={IEEE transactions on image processing},
  volume={13},
  number={4},
  pages={600--612},
  year={2004},
  publisher={IEEE}
}
\bibliographystyle{icml2025}



\end{document}